# Stress Testing Chain-of-Thought Prompting for Large Language Models

Aayush Mishra[1] and Karan Thakkar[2]

[1]Department of Computer Science, Johns Hopkins University
[2]Department of Electrical and Computer Engineering, Johns Hopkins University

September 28, 2023

## Abstract

This report examines the effectiveness of Chain-of-Thought (CoT) prompting in improving the multi-step reasoning abilities of large language models (LLMs). Inspired by previous studies [8], we analyze the impact of three types of CoT prompt perturbations, namely CoT order, CoT values, and CoT operators on the performance of GPT-3 on various tasks. Our findings show that incorrect CoT prompting leads to poor performance on accuracy metrics. Correct values in the CoT is crucial for predicting correct answers. Moreover, incorrect demonstrations, where the CoT operators or the CoT order are wrong, do not affect the performance as drastically when compared to the value based perturbations. This research deepens our understanding of CoT prompting and opens some new questions regarding the capability of LLMs to learn reasoning in context.

## 1 Introduction

Wei et. al. [11] recently showed that Chain-of-Thought (CoT) prompting (See Figure 1) significantly improves the performance of large language models on various numerical and reasoning tasks. Despite its success, there is little understanding of what makes CoT prompting effective in extracting correct answers from LLMs when prompted with a few demonstrations [2]. Recent findings also reveal that in-context learning could be very different from fine-tuning/training; for example, [8, 10] show that providing random labels or misleading instructions in context only marginally harms

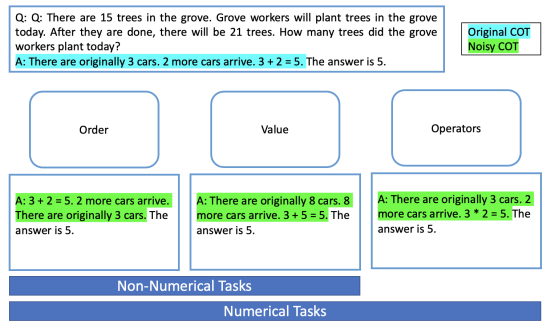

Figure 1: Examples describing the different perturbations examined in this study

model performance for certain tasks.

This raises the question of whether CoT prompting also shows a similar performance curve with incorrect prompts. As a pretest, we prompted a few sample problems with an incorrect CoT and found that the responses of an LLM still remained correct. To further explore how and why CoT prompting works, we designed a series of ablation experiments where we deliberately changed different aspects of the demonstrated CoTs, and measured the model performance accuracy on various tasks. Our hypothesis: the model only uses **structural context** from the provided CoT and doesn't necessarily depend on its logical correctness.

This is an important hypothesis to test because if true, we can drastically improve the performance of language models on various tasks without deliberate CoT prompting which requires human effort.

In section 2, we discuss the various prompting technique compared in our study. Section 3

Figure 2: Number of samples per dataset used to evaluate the LLM

describes our experimental setup in detail. The results of our experiments are then presented in section 4. Lastly, we discuss our conclusions and scope for future work in later sections.

## 2 Prompting LLMs

The three prompting techniques compared in our experiments are:

- **Few-Shot**: Provide a few questions along with their respective answer. This lets the model decide what reasoning to use for getting to the answer. Nothing other than the answer is provided in the prompt. This is the standard prompting technique used in extracting knowledge from LLMs.

- **CoT**: Provide a few questions along with their respective answer that follows from an explicitly described CoT. The CoT helps the model understand what reasoning to deduce from the prompt [11].

- **Perturbed CoT**: Provide the same prompts as in the case of CoT but with some perturbations that make the CoT logically incorrect while keeping the correct answer untouched. We define three types of perturbations. (refer Fig. 1)

    - **CoT values**: Changing the objective values in the CoT.
    - **CoT order**: Changing the order of sentences in the CoT.
    - **CoT operators**: Changing the operators (+, -, /, *) in CoT sentences (only applicable on numerical tasks).

Using the third novel category, we want to stress test the capability of CoT prompting. If the performance does not drop considerably with perturbed CoTs, it would be strong evidence in favor of our hypothesis. For this reason, we only perturb the CoT and keep the answers that follow them to be correct. With this experiment, we expect to gain a better understanding of how crucial the correctness of CoT is in performance gains.

## 3 Experimental Setup

### 3.1 Model

In this study, we only tested GPT-3. We used the default settings provided by the OpeanAI API with model=text-davinci-002, temperature=0.6 and max_tokens=1024.

### 3.2 Tasks and Datasets

Two main reasoning tasks are considered in the study: numerical and non-numerical. The datasets used for them are described briefly below.

- Numerical Tasks

    - **ASDiv** [7] a diverse English math word problem (MWP) corpus with 2,305 MWPs covering various language usage patterns and problem types.
    - **GSM8k** [3] a dataset of 8.5K high quality linguistically diverse grade school math word problems.
    - **SVAMP** [9] Modified versions of grade four or lower math problems to make MWPs more challenging.
    - **MAWPS** [5] - Multiple difficulty levels (single operator, single equation, adddition/subtraction only, and, multiple arithmetic operations)

In total, 7 numerical tasks, all math word problems of different styles and different levels of difficulty. An example question:

*Faye was cutting out some fabric for a friend. She cut a piece that was 3 centimeters wide and had an area of 24 square centimeters. How long was the piece?*

- Non-Numerical Tasks:
  - **Date Understanding** (from BIG Bench [1]). Tests a mix of common sense reasoning and a bit of numerical computation.
  - Symbolic reasoning [11]:
    * **Coin flip**. e.g.: *A coin is heads up. Christie flips the coin. Jaymie flips the coin. Is the coin still heads up?*
    * **Concat**. e.g.: *Take the last letters of the words in "Valentin Owen" and concatenate them.*
    * **Reverse list**. e.g.: *Reverse the sequence "umbrella, head, camera, battery, scissors".*

The number of samples used for evaluation per category (i.e. few-shot, CoT, and perturbed CoTs) is described in Fig 2. These numbers were decided based on budget constraints and the diversity of individual task.

### 3.3 Evaluation

After collecting responses for each category, accuracy for each task is calculated from the responses for performance evaluation. To get the final answer we parse the response and check for syntax similar to the examples given the prompting. If found, we take the final answer token or value as the prediction and match it with the ground truth labels in the dataset. As we keep the answer format (i.e "The answer is <prediction>") constant across the experiments we expect the model to do the same when predicting. If the answer format is violated, we consider it a false prediction irrespective of the value or token it predicts.

## 4 Results

In this section, we discuss the results of the experiments performed along with our observations. The following subsections discuss the results and observations for each group: Non-numerical and Numerical separately followed by general observations.

### 4.1 Non-Numerical Tasks

Fig 3 summarizes the performance of GPT-3 on the non-numerical reasoning tasks.

- **Reverse list** and **Concat**: We see that these two tasks are probably quite trivial for the model to understand as standard few-shot prompting itself works almost perfectly. However, we do see a drop in performance for *Order* and *Value* based perturbed CoT prompts. This suggests that *providing a wrong context can mislead the model* even when provided with the correct answer.

- **Date Understanding**: We see that almost all types of prompts have almost similar performance and CoT prompts works best. We also see a drop in performance when the CoT is perturbed.

- **Concat**: This case provides strong support for the effectiveness of correct CoT based prompting and how it can effectively turn the model from dense to expert. However, we don't understand why the performance varies so drastically for this particular task.

### 4.2 Numerical Tasks

Fig 4 summarizes the performance of GPT-3 on various numerical reasoning tasks.

- **MAWPS**: From Fig 4 (a), we see a general trend that perturbing numerical values have a strong negative impact on performance, which goes even below the standard few-shot prompting level. *This strongly suggests that the model can be misled or fooled using some value based adversarial examples.*

- **ASDiv, SVAMP**: We see a similar trend that CoT value perturbations are most severely hit in performance while correct CoT stays on top of the others (Fig. 4 (b)).

- **GSM**: This is more difficult compared to other tasks and we see that CoT helps a lot in comparison to standard few-shot prompting. We also see that even value perturbed CoT based prompts perform better than few-shot prompting in this case. (Fig. 4 (b))

### 4.3 General Observations

We noticed some peculiarities in the responses generated by GPT-3 and also some patterns in the accuracy plots.

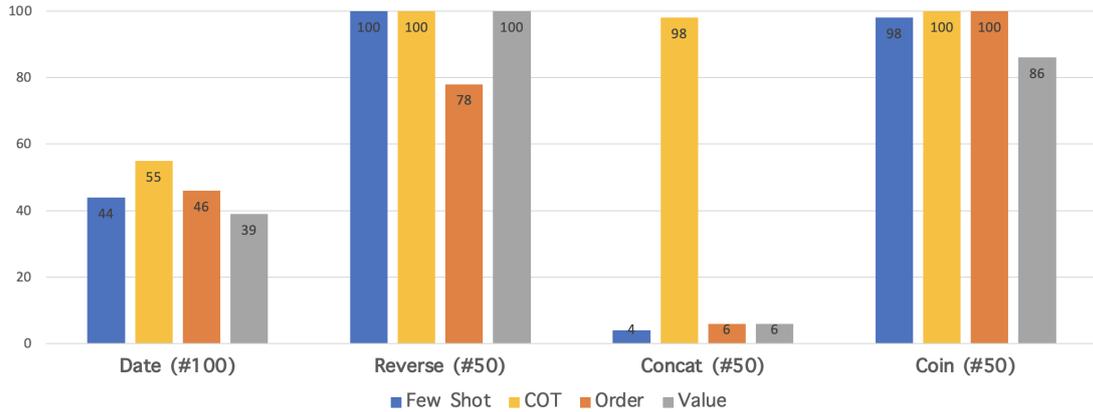

Figure 3: Performance of the LLM for Non-Numerical Tasks for different modes. '#' represents the number of samples per mode

- In the Date Understanding task, we saw several variations in the answer format generated by the model. For example, *format variation*: The model generated various date formats like Aug. 25th, 2021, 7/8/1972 vs 07/08/1972, etc. when it was explicitly asked to generate answers in the MM/DD/YYYY format. The model also generated *answer structure variation* like "... so tomorrow is 11/23/2001" vs "... answer is 11/23/2001" when all the prompts had the same answer structure. These variations did not show up in other tasks and datasets. *This suggests that the model probably has poor understanding of date format templates.*

- *Larger the gap in performance between few-shot vs CoT, implies better performance retention after CoT perturbation.* We see a general trend in the numerical reasoning tasks that in cases where CoT prompting doesn't drastically improve performance over few-shot prompting, perturbing the CoT has a severe impact on performance. However, in cases like GSM and MAWPS (Multi-Arith), the performance retention is higher and mostly better than few-shot. Interestingly, these are also the toughest tasks in terms of the number of steps to be performed to get to the answer. *This suggests that the model might not depend as much on the CoT correctness in tougher tasks, as it does on contextual awareness provided by even the incorrect CoT.*

- *Retention of CoT Perturbation*: Although we only quantitatively test on the correctness of the answer generated by the model, we also noticed many cases where the model retains the incorrect CoT structure while coming up with the correct answer. This is particularly surprising because to do this systematically, the model must understand:

  1. What is asked in the question and come up with correct reasoning to derive it.
  2. What is the pattern in the CoT prompts and generate that pattern independent of how the answer is generated.

An assortment of such responses generated by the model is provided below.

**Wrong CoT order**
*Q: Today is the last day of the first quarter of 2008. What is the date one week ago from today in MM/DD/YYYY?*
*A: So one week ago from today was 03/24/2008. Today is 03/31/2008, which is the last day of the first quarter. So the answer is 03/24/2008.*

**Wrong CoT value/operator**
*Q: Isabel uploaded 2 pictures from her phone and 4 from her camera to Facebook. If she sorted the pics into 3 different albums with the same amount of pics in each album, how many pictures were in each of the albums?*
*A: Isabel had 18 pictures in total. 2 + 4 = 8. 18 / 8 is 2. So each album would have 2 pictures.*

## 5 Conclusions

It is straightforward to see from our results that the correctness of Chain-of-Thought prompts is essential to performance gains when compared to standard few-shot prompting. This is unlike [8] which suggested that the correctness of few-shot samples might not matter as much as we might think. In general, we found out that the correctness of the *values* in CoT is most critical to performance. Perturbing them might mislead the model and performance may drop even below standard few-shot levels. Our experiments reveal that our hypothesis is wrong and we can not skip creating correct CoTs for performance gains. Although the results of our experiments provide support for CoT prompting, there might still be better/easier methods to extract performance out of LLMs that do not require manual labeling efforts [4, 12, 6].

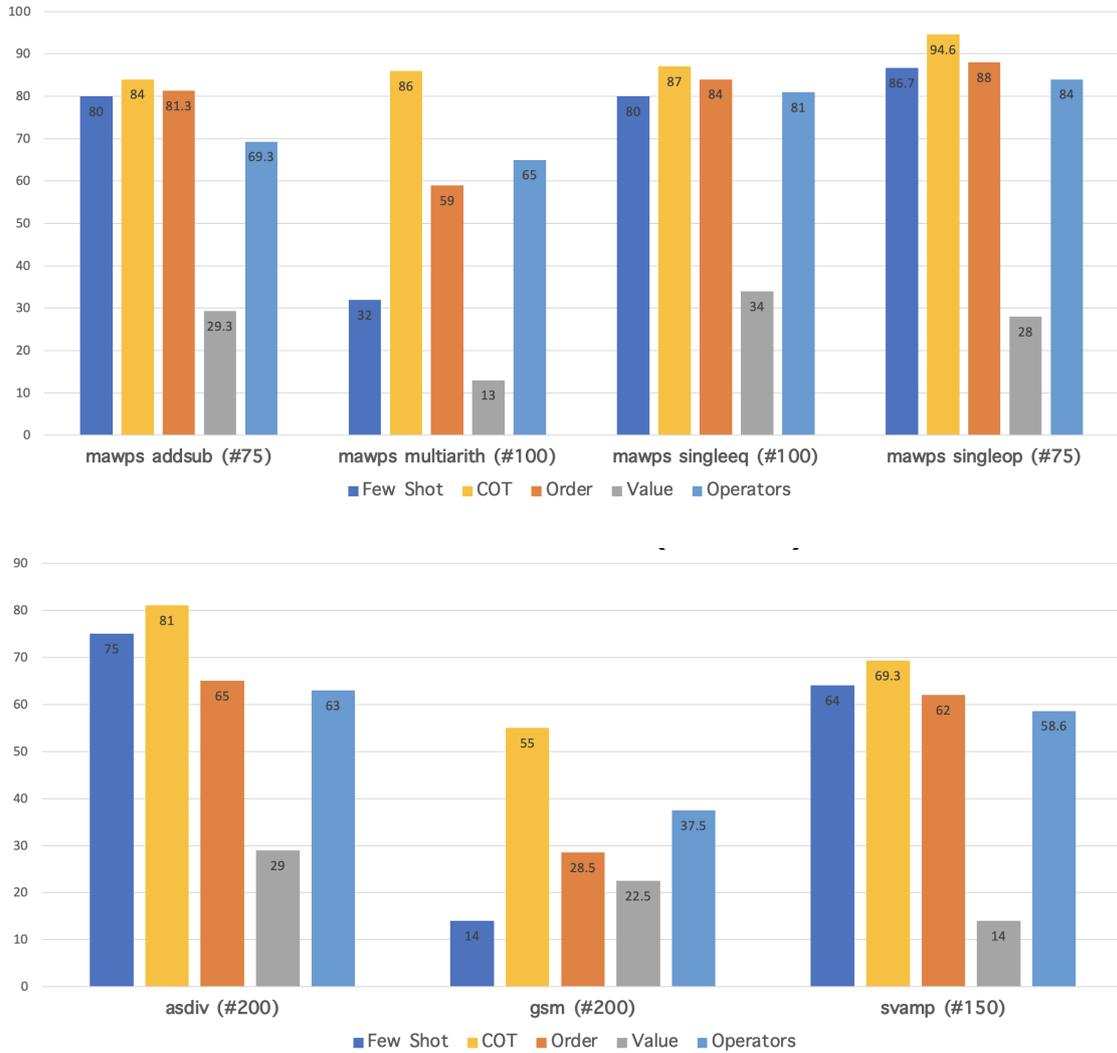

Figure 4: (a) top: Performance of the LLM for Numerical Tasks for different modes on the MAWPS dataset. (b) bottom: Performance of the LLM for Numerical Tasks for different modes on ASDIV, GSM, and SVAMP datasets.

## 6  Future Directions

Our experiments reveal several interesting phenomena which spawn potential future works.

- **Adversarial Examples**: We saw that value-based perturbations severely hit performance, which even falls below standard few-shot prompting in most cases. Although we changed each quantitative value in the CoT for this type of perturbation, there might be scope for changing just a few or even a single value in a prompt which could result in a severe performance hit. If true, this would mean that LLMs suffer from the same adversarial sample problems as standard neural networks.

- **Retention of CoT Perturbation** is an interesting by-product that we noticed in many model responses. This motivates a study to systematically identify what is causing the models to retain the perturbed CoT structure while coming up with correct answers.

- **Difficulty of task vs correctness of CoT**: We observed that for more difficult problems, even perturbed CoTs provide some benefits over standard prompting. Is the structural context provided by CoTs all that is being used in those cases? This motivates future studies on what exactly is the recipe for the perfect prompt.

## 7  Contribution

- **Aayush Mishra**: Experiment design, coding perturbations.

- **Karan Thakkar**: Experiment design, collecting results.

- **Shared**: Brainstorming ideas, presentations, and report.

## Acknowledgements

We would like to thank Professor Daniel Khashabi (danielk@cs.jhu.edu) for the constant support and necessary resources required for the completion of this project. We would also like to thank our classmates for their valuable feedback during project presentations.